%% file: naaclhlt2019.tex
\documentclass[11pt,a4paper]{article}
\usepackage[hyperref]{naaclhlt2019}
\usepackage{times}
\usepackage{latexsym}
\usepackage{braket}
\usepackage{helvet}
\usepackage{courier}
\usepackage{amsfonts}
\usepackage{multirow,array}
\usepackage{natbib}
\usepackage{url}
\usepackage{amssymb,amsmath}
\usepackage{booktabs}
\usepackage{mathrsfs}
\usepackage{makecell}
\usepackage[ruled,vlined]{algorithm2e}
\usepackage{graphicx}
\usepackage{subcaption}
\usepackage[toc,page]{appendix}

\usepackage{color,xcolor}
\usepackage{todonotes}

\aclfinalcopy 
\def\aclpaperid{440} 

\newcommand*\samethanks[1][\value{footnote}]{\footnotemark[#1]}

\title{CNM: An Interpretable Complex-valued Network for Matching}

\author{Qiuchi Li\Thanks{Equal Contribution} \\
  University of Padua  \\
  Padua, Italy \\
  {\tt qiuchili@dei.unipd.it} \\
  \And
  Benyou Wang \samethanks\\
   University of Padua  \\
  Padua, Italy \\
  {\tt wang@dei.unipd.it} \\
  \And
  Massimo Melucci\Thanks{Corresponding Author} \\
  University of Padua  \\
  Padua, Italy \\
  {\tt melo@dei.unipd.it} \\}

\date{}

\begin{document}
\maketitle
\begin{abstract}
This paper seeks to model human language by the mathematical framework of quantum physics. With the well-designed mathematical formulations in quantum physics, this framework unifies different linguistic units in a single complex-valued vector space, e.g. words as particles in quantum states and sentences as mixed systems. A complex-valued network is built to implement this framework for semantic matching. With well-constrained complex-valued components, the network admits interpretations to explicit physical meanings. The proposed complex-valued network for matching (CNM)\footnote{https://github.com/wabyking/qnn.git} achieves comparable performances to strong CNN and RNN baselines on two benchmarking question answering (QA) datasets. 
\end{abstract}



\input{introduction.tex}

\input{background.tex}
\input{methodology.tex}

\input{experiment.tex}

\input{discussion.tex}
\input{conclusion.tex}
\input{acknowledge.tex}

\bibliographystyle{acl_natbib}
\bibliography{naacl2019}

\end{document}

%% file: introduction.tex
\section{Introduction}
\label{sec:introduction}
There is a growing concern on the interpretability of neural networks. Along with the increasing power of neural networks comes the challenge of interpreting the numerical representation of network components into human-understandable language. ~\citet{lipton_mythos_2018} points out two important factors for a model to be interpretable, namely \emph{post-hoc interpretability} and \emph{transparency}. The former refers to explanations of why a model works after it is executed, while the latter concerns self-explainability of components through some mechanisms in the designing phase of the model.  

We seek inspirations from quantum physics to build transparent and post-hoc interpretable networks for modeling human language. The emerging research field of cognition suggests that there exist quantum-like phenomena in human cognition~\citep{aerts_quantum_2014}, especially language understanding~\cite{bruza2008entangling}. Intuitively, a sentence can be treated as a physical system with multiple words (like particles), and these words are usually polysemous (superposed) and correlated (entangled) with each other. Motivated by these existing works, we aim to investigate the following Research Question (RQ).

\textit{ \textbf{RQ1}: Is it possible to model human language with the mathematical framework of quantum physics?}

Towards this question, we build a novel quantum-theoretic framework for modeling language, in an attempt to capture the quantumness in the cognitive aspect of human language. The framework models different linguistic units as quantum states with adoption of quantum probability (QP), which is the mathematical framework of quantum physics that models uncertainly on a uniform \textit{Semantic Hilbert Space} (SHS). 

Complex values are crucial in the mathematical framework of characterizing quantum physics. In order to preserve physical properties, the linguistic units have to be represented as complex vectors or matrices. This naturally gives rise to another research question:

 \textit{ \textbf{RQ2}: Can we benefit from the complex-valued representation of human language in a real natural language processing (NLP) scenario? }
 
To this end, we formulate a linguistic unit as a complex-valued vector, and link its length and direction to different physical meanings: the length represents the relative weight of the word while the direction is viewed as a superposition state. The superposition state is further represented in an amplitude-phase manner, with amplitudes corresponding to the lexical meaning and phases implicitly reflecting the higher-level semantic aspects such as polarity, ambiguity or emotion.

In order to evaluate the above framework, we implement it as a complex-valued network (CNM) for  semantic matching. The network is applied to the question answering task, which is the most typical matching task that aims at selecting the best answer for a question from a pool of candidates. 
In order to facilitate local matching with n-grams of a sentence pair, we design a local matching scheme in CNM. Most of State-of-the-art QA models are mainly based on Convolution Neural Network (CNN), Recurrent Neural Network (RNN) and many variants thereof~\citep{wang_long_2015,yang_anmm:_2016,hu_convolutional_2014,tan_lstm-based_2016}. However, with opaque structures of convolutional kernels and recurrent cells, these models are hard to understand for humans. We argue that our model is advantageous in terms of interpretability.
 
Our proposed CNM is transparent in that it is designed in alignment with quantum physics. Experiments on benchmarking QA datasets show that CNM has comparable performance to strong CNN and RNN baselines, whilst admitting post-hoc interpretations to human-understandable language. We therefore answer RQ1 by claiming that it is possible to model human language with the proposed quantum-theoretical framework in this paper. Furthermore, an ablation study shows that the complex-valued word embedding performs better than its real counterpart, which allows us to answer RQ2 by claiming that we benefit from the complex-valued representation of natural language on the QA task.

%% file: background.tex
\section{Background}
Here we briefly introduce quantum probability and discuss a relevant work on quantum-inspired framework for QA.

\subsection{Quantum Probability}


Quantum probability provides a sound explanation for the phenomena and concepts of quantum mechanics, by formulating events as subspaces in a vector space with projective geometry. 

\subsubsection{Quantum Superposition}
Quantum Superposition is one of the fundamental concepts in Quantum Physics, which describes the uncertainty of a single particle. In the micro world, a particle like a photon can be in multiple mutual-exclusive basis states simultaneously with a probability distribution. In a two-dimensional example, two basis vectors are denoted as $\ket{0}$ and $\ket{1}$\footnote{We here adopt the widely used Dirac notations in quantum probability, in which a \textit{unit} vector $\vec{\mu}$ and its transpose $\vec{\mu}^T$ are denoted as a ket $\vert u \rangle$ and a bra $\langle u \vert$ respectively. }. \emph{Superposition} is implemented to model a  general state which is a linear combination of basis vectors with complex-valued weights such that 
\begin{equation}   
\label{eq:superpostion}
\ket{\phi} = \alpha_0 \ket{0} + \alpha_1 \ket{1}, 
\end{equation} 
where $\alpha_0$ and $\alpha_1$ are complex scalars satisfying $0 \leq |\alpha_0|^2 \leq 1 $, $0 \leq |\alpha_1|^2 \leq 1 $  and $ |\alpha_0|^2 + |\alpha_1|^2 = 1$. It follows that $\ket{\phi}$ is  defined over the complex field. When $\alpha_0$ and $\alpha_1$ are non-zero values, the state $\ket{\phi}$ is said to be a superposition of the states $\ket{0}$ and $\ket{1}$, and the scalars $\alpha_0$ and $\alpha_1$ denote the probability amplitudes of the superposition.

\subsubsection{Measurement}
The uncertainty of an ensemble system with multiple particles is encapsulated as a mixed state, represented by a positive semi-definite matrix with unitary trace called \emph{density matrix}: $ \rho = \sum_i^m \ket{\phi_i} \bra{\phi_i} $, where $\{\ket{\phi_i}\}_{i=0}^m$ are pure states like Eq.~\ref{eq:superpostion}.  In order to infer the probabilistic properties of $\rho$ in the state space, Gleason's theorem~\cite{gleason1957measures,hughes1992structure} is used to calculate probability to observe $x$ through projection measurements $\ket{x}\bra{x}$ that is a rank-one projector denoted as a outer product of $\ket{x}$.
\begin{equation}
\label{trace_measurement}
p_x(\rho) = \bra{x} \rho \ket{x} = tr(\rho \ket{x}\bra{x})
\end{equation} 
The measured probability $p_x(\rho)$ is a non-negative real-valued scalar, since both $\rho$ and $\ket{x} \bra{x}$ are Hermitian. The unitary trace property guarantees $\sum_{x \in X} p_x(\rho) = 1$ for $X$ being a set of orthogonal basis states.




\subsection{Neural Network based Quantum-like Language Model (NNQLM)}
Based on the density matrices representation for documents in information retrieval~\cite{van2004geometry,sordoni2013modeling}, ~\citet{zhang_end_to_end_2018} built a neural network with density matrix for question answering. This Neural Network based Quantum Language Model (NNQLM) embeds a word as a unit vector and a sentence as a real-valued density matrix. The distance between a pair of density matrices is obtained by extracting features of their matrix multiplication in two ways: NNQLM-I directly takes the trace of the resulting matrix, while NNQLM-II applies convolutional structures on top of the matrix to determine whether the pair of sentences match or not.

NNQLM is limited in that it does not make proper use of the full potential of probabilistic property of a density matrices.
By treating density matrices as ordinary real vectors (NNQLM-I) or matrices (NNQLM-II), the full potential with complex-valued formulations is largely ignored. Meanwhile, adding convolutional layers on top of a density matrix is more of an empirical workaround than an implementation of a theoretical framework.

In contrast, a complex-valued matching network is built on top of a quantum-theoretical framework for natural language. In particular, an indirect way to measure the distance between two density matrices through trainable measurement operations, which makes advantage of the probabilistic properties of density matrices and also provides flexible matching score driven by training data.

%% file: methodology.tex
\section{Semantic Hilbert Space}



\label{semantic_hilbert_space}
Here we introduce the Semantic Hilbert Space $\mathcal{H}$  defined on a complex vector space $\mathcal{C}^n$, and three different linguistic units, namely sememes, words and word combinations on the space. The concept of semantic measurement is introduced at last.



\textbf{Sememes.} 
We assume $\mathcal{H}$ is spanned by the set of orthogonal basis states $\{\ket{e_j}\}_{j=1}^n$ for \textit{sememes}, which are the minimum semantic units of word meanings in language universals~\cite{goddard1994semantic}. The unit state $\ket{e_j}$ can be seen as a one-hot vector, i.e., the $j$-th element in $\ket{e_j}$ is one while other elements are zero, in order to obtain a set of orthogonal unit states. Semantic units with larger granularities are based on the set of sememe basis. 

\textbf{Words.} 
Words are composed of sememes in superposition. Each word $w$ is a superposition over all sememes $\{\ket{e_j}\}_{j=1}^n$, or equivalently a unit-length vector on $\mathcal{H}$:
\begin{equation}
\label{superposition}
\vert w \rangle =  \sum_{j=1}^n r_j e^{i\phi_j} \vert e_j \rangle,
\end{equation}
$i$ is the imaginary number with $i^2=-1$. In the above expression, $\{r_j\}_{j=1}^n$ are non-negative real-valued amplitudes satisfying $\sum_{j=1}^n {r_j}^2$ =1 and $\phi_j \in [-\pi,\pi]$ are the corresponding complex phases. In comparison to Eq.~\ref{eq:superpostion}, $\{r_j e^{i\phi_j}\}_{j=0}^n$ are the polar form representation of the complex-valued scalars $\{\alpha_j\}_{j=0}^1$.




\textbf{Word Combinations.} 
We view a combination of words (e.g. phrase, $n$-gram, sentence or document) as a mixed system composed of individual words, and its representation is computed as follows:
\begin{equation}
\label{mixture_old}
\rho = \sum^m_j { \frac{1}{m} \ket{w_j} \bra{w_j}},
\end{equation}
where $m$ is the number of words and $\ket{w_j}$ is word superposition state in Eq.~\ref{superposition}, allowing multiple occurrences. Eq.~\ref{mixture_old} produces a density matrix $\rho$ for semantic composition of words. It also describes a non-classical distribution over the set of sememes:  the complex-valued off--diagonal elements describes the correlations between sememes, while the diagonal entries (guaranteed to be real by its original property) correspond to a standard probability distribution. The off--diagonal elements provide our framework some potentials to model the possible interactions between the basic sememe basis, which was usually considered mutually independent with each other.


\textbf{Semantic Measurements.} 
The high-level features of a sequence of words are extracted through
measurements on its mixed state. Given a density matrix $\rho$ of a mixed state, a rank-one projector $P$, which is the outer product of a unit complex vector, i.e. $P=\ket{x}\bra{x}$, is applied as a measurement projector. It is worth mentioning that $\ket{x}$ could be any pure state in this Hilbert space (not only limited to a specific word $w$).
The measured probability is computed by Gleason`s Theorem in Eq.~\ref{trace_measurement}.

\section{Complex-valued Network for Matching}

We implemented an end-to-end network for matching on the Semantic Hilbert Space. Fig. 1 shows the overall structure of the proposed Complex-valued Network for Matching (CNM). Each component of the network is further discussed in this section. 
\begin{figure*}[t]\label{architecture}
\centering
  \caption{Architecture of Complex-valued Network for Matching. $\raisebox{.5pt}{\textcircled{\raisebox{-.9pt} {M}}} $ means a measurement operation according to Eq.~\ref{trace_measurement}.}
\includegraphics[width=0.9\textwidth]{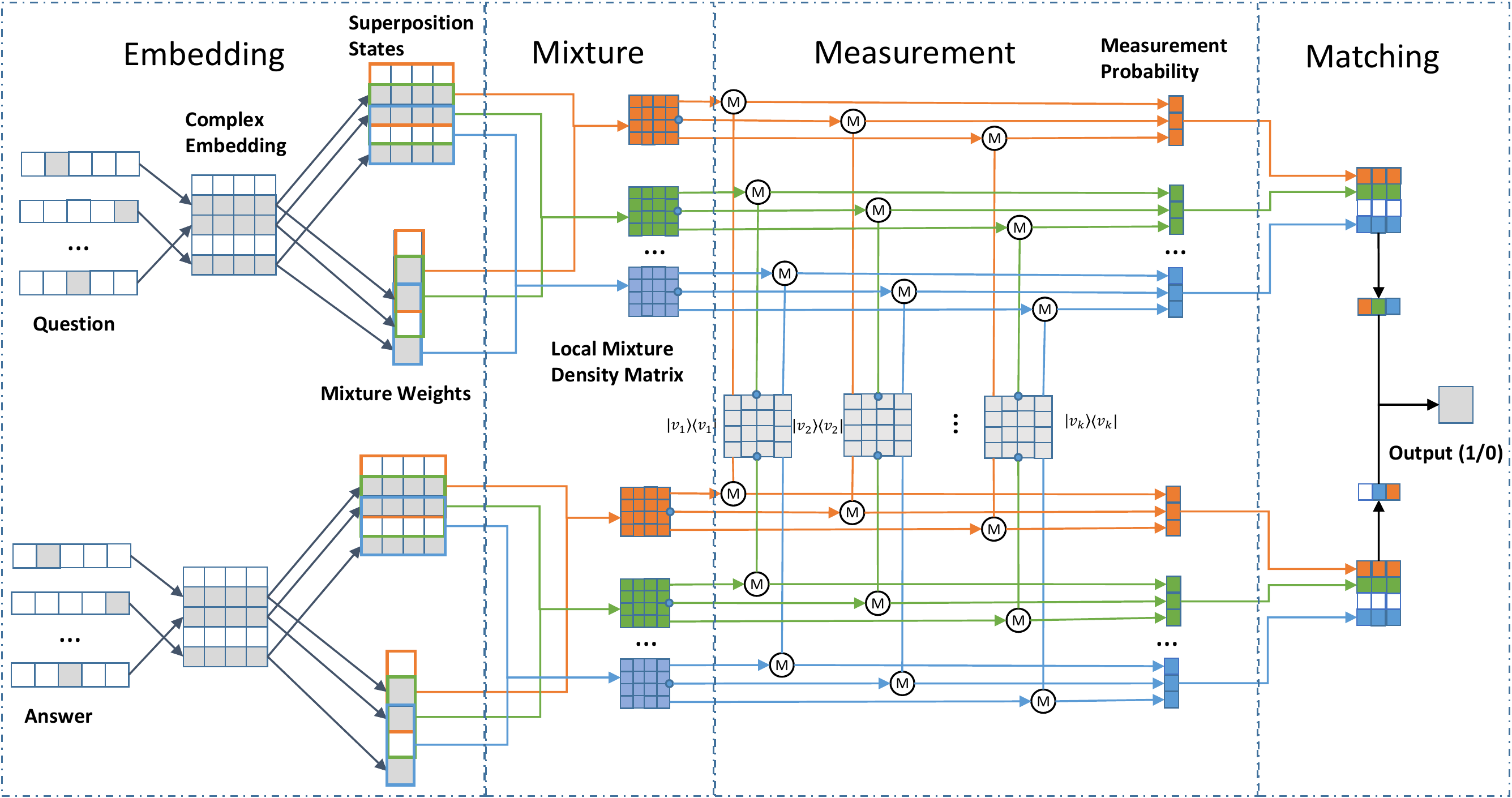}
\vspace{-10pt}
\end{figure*} 

\subsection{Complex-valued Embedding}
On the Semantic Hilbert Space, each word $w$ is embedded as a complex-valued vector $\vec{w}$. Here we link its length and direction to different physical meanings: the length of a vector represents the relative weight of the word while the vector direction is viewed as a superposition state. Each word $w$ adopts a  normalization into a superposition state $\ket{w}$ and a word-dependent weight $\pi(w)$:
\begin{eqnarray}
\label{normalization}
\ket{w} = \frac{\vec{w}}{||\vec{w}||}, ~~\pi(w) = ||\vec{w}||,
\end{eqnarray}
where $||\vec{w}||$ denotes the 2-norm length of $\vec{w}$. $\pi(w)$ is used to compute the relative weight of a word in a local context window, which we will elaborate in Section 4.2. 




\subsection{Sentence Modeling with Local Mixture Scheme}

A sentence is modeled as a combination of individual words in it. NNQLM~\cite{zhang_end_to_end_2018} models a sentence as a~\textit{global mixture} of all words, which implicitly assumes a global interaction among all sentence words. This seems to be unreasonable in practice, especially for a long text segment such as a paragraph or a document, where the interaction between the first word and the last word is often negligible. Therefore, we address this limitation by proposing a \textit{local mixture} of words, which tends to capture the semantic relations between neighbouring words and undermine the long-range word dependencies.  As is shown in Fig. 2, a sliding window is applied and a density matrix is constructed for a local window of length $l$ (e.g. 3). Therefore, a sentence is composed of a sequence of density matrices for $l$-grams.


\begin{figure}[t]\label{fig:mixture}
  \caption{Architecture of local mixture component. A sliding window in black color is applied to the sentence, generating a local mixture density matrix for each local window of length $l$. $\bigodot$ means that a matrix multiplies a number with each elements. $\bigotimes$ denotes an outer product of a vector.}
\includegraphics[width=77mm]{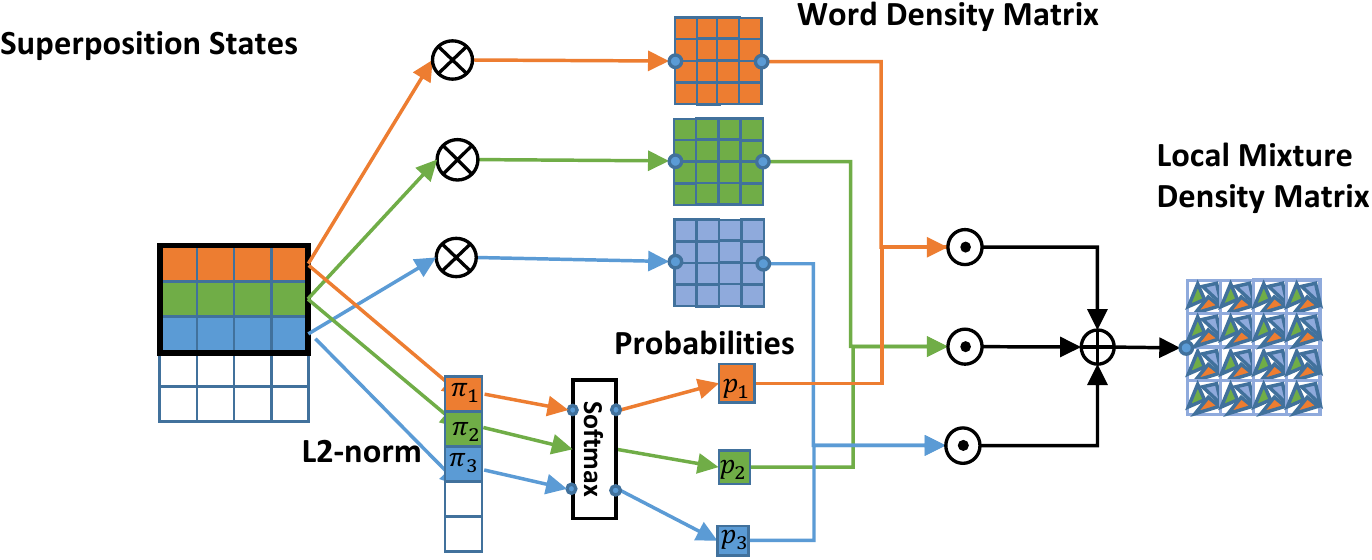}
\vspace{-10pt}
\end{figure}

The representation of a local $l$-gram window is obtained by an improved approach over Eq.~\ref{mixture_old}. In Eq.~\ref{mixture_old}, each word is assigned with the same weight, which does not hold from an empirical point of view. In this study, we take the $L2$-norm of the word vector as the relative weight in a local context window for a specific word, which could be updated during training. To some extent, $L2$-norm is a measure of semantic richness of a word, i.e. the longer the vector the richer the meaning. The density matrix of an $l$-gram is computed as follows:
 

\begin{equation}\label{mixture}
\rho = \sum^l_i { p(w_i) \ket{w_i} \bra{w_i}},
\end{equation}
where the relative importance of each word $p(w_i)$ in an $l$-gram is the soft-max normalized word-dependent weight: $\label{softmax} p(w_i) = \frac{e^{\pi(w_i)}}{\sum^l_j e^{\pi(w_j)}}$, where $\pi(w_i)$ is the word-dependent weight. By converting word-dependent weights to a probability distribution, a legal density matrix is produced, because $\sum^l_i { p(w_i)} =1$ gives $tr(\rho) = 1$. Moreover, the weight of a word also depends on its neighboring words in a local context.

\subsection{Matching of Question and Answer}
In quantum information, there have been works trying to estimate a quantum state from the results of a series of measurements~\citep{PhysRevA.63.040303,lvovsky2004}. Inspired by these works, we introduce trainable measurements to extract density matrix features and match a pair of sentences. 

Suppose a pair of sentences with length $L$ are represented as two sets of density matrices $\{\rho_{1j}\}_{j=1}^L$ and $\{\rho_{2j}\}_{j=1}^L$ respectively. The same set of $K$ semantic measurement operators $\{\ket{v_k}\}_{k=1}^K$ are applied to both sets, producing a pair of $k$-by-$L$ probability matrix $p^1$ and $p^2$, where $p_{jk}^1 = \bra{v_k} \rho_{1j} \ket{v_k}$ and $p_{jk}^2 = \bra{v_k} \rho_{2j} \ket{v_k}$ for $k \in \{1,...,K\}$ and $j \in \{1,...,L\}$. A classical vector-based distances between $p^1$ and $p^2$ can be computed as the matching score of the sentence pair. By involving a set of semantic measurements, the properties of density matrix are taken into consideration in computing the density matrix distance. 

We believe that this way of computing density matrix distance is both theoretically sound and applicable in practice. The trace inner product of density matrices~\cite{zhang_end_to_end_2018} breaks the basic axioms of metric, namely non-negativity, identity of indiscernables and triangle inequality. The CNN-based feature extraction~\cite{zhang_end_to_end_2018} for density matrix multiplication loses the property of density matrix as a probability distribution.~\citet{nielsen_quantum_2010} introduced three measures namely trace distance, fidelity and VN-divergence. However, it is computationally costly to compute these metrics and propagate the loss in an end-to-end training framework. 

We set the measurements to be trainable so that the matching of question and answering can be integrated into the whole neural network, and identify the discriminative semantic measurements in a data-driven manner. From the perspective of linear discriminant analysis (LDA)~\cite{fisher_use_1936}, this approach is intended to find a group of finite discriminative projection directions for a better division of different classes, but in a more sound framework inspired by quantum probability with complex-valued values. From an empirical point of view, the data-driven measurements make it flexible to match two sentences.

%% file: experiment.tex
\section{Experiments}
\subsection{Datasets and Evaluation Metrics}
The experiments were conducted on two benchmarking question answering datasets for question answering (QA), namely TREC QA~\citep{voorheesellenmticedawnm.2000} and WikiQA~\citep{yang_wikiqa:_2015}. TREC QA is a standard QA dataset in the Text REtrieval Conference (TREC). WikiQA is released by Microsoft Research on open domain question answering. On both datasets, the task is to select the most appropriate answer from the candidate answers for a question, which require a ranking of candidate answers. After removing the questions with no correct answers, the statistics of the cleaned datasets are given in the Tab.~\ref{table:statistic}. Two common rank-based metrics, namely mean average precision (MAP) and mean reciprocal rank (MRR), are used to measure the performance of models.

\begin{table}[t]
\begin{center}
\small
\caption{Dataset Statistics. For each cell, the values denote the number of questions and question-answer pairs respectively.}
\label{table:statistic}
\begin{tabular}{l|l|l|l}
 \hline
  Dataset & train & dev & test  \\ \hline
 TREC QA & 1229/53417 & 65/117   & 68/1442   \\
 WikiQA &873/8627 &126/130  &633/2351\\
 \hline
\end{tabular}
\end{center}
\vspace{-15pt}
\end{table}

\subsection{Experiment Details}

\subsubsection{Baselines}
We conduct a comprehensive comparison across a wide range of models. On TREC QA the experimented models include Bigram-CNN~\citep{yu_deep_2014}, three-layered Long Short-term Memory (LSTM) in combination with BM25 (LSTM-3L-BM25)~\citep{wang_long_2015}, attention-based neural matching model (aNMM)~\citep{yang_anmm:_2016}, Multi-perspective CNN (MP-CNN)~\citep{he_multi-perspective_2015}, CNTN~\cite{qiu2015convolutional}, attention-based LSTM+CNN model (LSTM-CNN-attn)~\citep{tang_document_2015} and pairwise word interaction modeling (PWIM)~\citep{he_pairwise_2016}. On WikiQA dataset, we involve the following models into comparison: Bigram-CNN~\citep{yu_deep_2014}, CNN with word count information (CNN-Cnt)~\citep{yang_wikiqa:_2015}, QA-BILSTM \cite{santos2016attentive},  BILSTM with attentive pooling (AP-BILSTM)~\cite{santos2016attentive},
and LSTM with attention (LSTM-attn)~\citep{miao_neural_2015}. On both datasets, we report the results of quantum language model~\citep{sordoni2013modeling} and two models NNQLM-I, NNQLM-II by~\citep{zhang_end_to_end_2018} for comparison.

\subsubsection{Parameter Settings}

The parameters in the network are $\Theta = \{R, \Phi, \{\ket{v_i}\}_{i=1}^k\}$, in which $R$ and $\Phi$ denote the lookup tables for amplitudes and complex phases of each word, and $\ket{v_i}\}_{i=1}^k$ denotes the set of semantic measurements. We use 50-dimension complex word embedding.The amplitudes are initialized with 50-dimension Glove vectors~\citep{pennington2014glove} and L2-norm regularized during training. The phases are randomly initialized under a normal distribution of $[-\pi,\pi]$. The semantic measurements $\{\ket{v_i}\}_{i=1}^k\}$ are initialized with orthogonal real-valued one-hot vectors, and each measurement is constrained to be of unit length during training. We perform max pooling over the sentence dimension on the measurement probability matrices, resulting in a $k$-dim vector for both a question and an answer. We concatenate the vectors for $l =1,2,3,4$ for questions and answers, and larger size of windows are also tried. We will use a longer sliding window in datasets with longer sentences. The cosine similarity is used as the distance metric of measured probabilities. We use triplet hinge loss and set the margin  $\alpha = 0.1$. A dropout layer is built over the embedding layer and measurement probabilities with a dropout rate of 0.9.

A grid search is conducted over the parameter pools to explore the best parameters. The parameters under exploration include \{0.01,0.05,0.1\} for the learning rate, \{1e-5,1e-6,1e-7,1e-8\} for the L2-normalization of complex word embeddings, \{8,16,32\} for batch size, and \{50,100,300,500\} for the number of semantic measurements.

\subsubsection{Parameter Scale}
The proposed CNM has a limited scale of parameters. Apart from the complex word embeddings which are $|V| \times 2n$ by size, the only set of parameters are $\{\ket{v_i}\}_{i=1}^k$ which is $k\times 2n$, with $|V|, k, n$ being the vocabulary size, number of semantic measurements and the embedding dimension, respectively. In comparison, a single-layered CNN has at least $l\times k \times n$ additional parameters with $l$ being the filter width, while a single-layered LSTM is $4 \times k \times (k+n)$ by the minimum parameter scale.  Although we use both amplitude part and phase part for word embedding, lower dimension of embedding are adopted, namely 50, with the comparable performance.
Therefore, our network scales better than the advanced models on the CNN or LSTM basis.


\subsection{Experiment Results}

\begin{table}[t]
\begin{center}
\small
\caption{Experiment Results on TREC QA Dataset. The best performed values are in bold. }
\label{fig:trecqa}
\begin{tabular}{l|cc}
 \hline
Model & MAP & MRR \\ \hline
 

 Bigram-CNN & 0.5476&0.6437 \\
 LSTM-3L-BM25 & 0.7134 & 0.7913 \\
 LSTM-CNN-attn & 0.7279 & 0.8322 \\
 aNMM & 0.7495 & 0.8109 \\
 MP-CNN & \textbf{0.7770} & 0.8360 \\
 CNTN & 0.7278 &0.7831 \\
 PWIM & 0.7588 & 0.8219 \\
 \hline
  QLM & 0.6780 & 0.7260\\
 NNQLM-I & 0.6791 &0.7529  \\ 
 NNQLM-II & 0.7589 &0.8254 \\
 \hline
CNM & 0.7701 &\textbf{0.8591} \\Over NNQLM-II & 1.48\% $\uparrow$ &4.08\%  $\uparrow$ \\

 \hline

\end{tabular}
\end{center}
\vspace{-5pt}
\end{table}

\begin{table}[t]
\begin{center}

\small
\caption{Experiment Results on WikiQA Dataset.The best performed values for each dataset are in bold.}
\label{fig:wikiqa}
\begin{tabular}{l|cc}

 \hline
Model & MAP & MRR \\ \hline
 
 Bigram-CNN & 0.6190 & 0.6281 \\ 
QA-BILSTM& 0.6557 & 0.6695\\
AP-BILSTM& 0.6705 & 0.6842\\
LSTM-attn & 0.6639 & 0.6828 \\ 
CNN-Cnt & 0.6520 &  0.6652 \\

 \hline
 QLM & 0.5120 &  0.5150\\
 NNQLM-I & 0.5462 & 0.5574 \\
 NNQLM-II & 0.6496 & 0.6594 \\
\hline
 CNM & \textbf{0.6748} & \textbf{0.6864} \\ 
Over NNQLM-II & 3.88\% $\uparrow$ &4.09\%  $\uparrow$ \\
\hline

\end{tabular}
\end{center}
\vspace{-5pt}
\end{table}






Tab.~\ref{fig:trecqa} and \ref{fig:wikiqa} show the experiment results on TREC QA and WikiQA respectively, where bold values are the best performances out of all models. Our model achieves 3 best performances out of the 4 metrics on TREC QA and WikiQA, and performs slightly worse than the best-performed models on the remaining metric. This illustrates the effectiveness of our proposed model from a general perspective.

Specifically, CNM outperforms most CNN and LSTM-based models, which have more complicated structures and a relatively larger parameters scale. Also, CNM performs better than existing quantum-inspired QA models, QLM and NNQLM on both datasets, which means that the quantum theoretical framework gives rise to better performs model. Moreover, a significant improvement over NNQLM-1 is observed on these two datasets, supporting our claim that trace inner product is not an effective distance metric of two density matrices. 




\subsection{Ablation Test}
\label{sec:ablation}
An ablation test is conducted to examine the influence of each component on our proposed CNM. The following models are implemented in the ablation test. \emph{FastText-MaxPool} adopt max pooling over word-embedding, just like FastText~\citep{joulin2016bag}.
\emph{CNM-Real} replaces word embeddings and measurements with their real counterparts. \emph{CNM-Global-Mixture} adopts a global mixture of the whole sentence, in which a sentence is represented as a single density matrix, leading to a probability vector for the measurement result. \emph{CNM-trace-inner-product} replaces the trainable measurements with trace inner product like NNQLM.

For the real-valued models, we replace the embedding with double size of dimension, in order to eliminate the impact of the parameter scale on the performance. Due to limited space, we only report the ablation test result on TREC QA, and WikiQA has the similar trends. The test results in Tab.~\ref{table:Ablation} demonstrate that each component plays a crucial role in the CNM model. In particular, the comparison with CNM-Real and FastText-MaxPool shows the effectiveness of introducing complex-valued components, the increase in performance over CNM-Global-Mixture reveals the superiority of local mixture, and the comparison with CNM-trace-inner-product confirms the usefulness of trainable measurements.

\begin{table}[t]
\begin{center}
\small
\caption{Ablation Test. The values in parenthesis are the performance difference between the model and CNM.}\label{table:Ablation}

\resizebox{0.48\textwidth}{!}{
\begin{tabular}{lll}
 \hline
  Setting &MAP  & MRR\\ \hline
FastText-MaxPool & 0.6659~(0.1042$\downarrow$) & 0.7152~(0.1439$\downarrow$)\\ 
CNM-Real & 0.7112~(0.0589$\downarrow$) & 0.7922~(0.0659$\downarrow$)\\ 
CNM-Global-Mixture & 0.6968~(0.0733$\downarrow$) & 0.7829~(0.0762$\downarrow$)  \\ 
CNM-trace-inner-product & 0.6952~(0.0749$\downarrow$) & 0.7688~(0.0903$\downarrow$) \\
\hline
CNM & 0.7701 & 0.8591  \\
 \hline
\end{tabular}
}
\end{center}
\vspace{-10pt}
\end{table}

%% file: discussion.tex
\section{Discussions}
\label{discussion}

This section aims to investigate the proposed research questions mentioned in Sec~\ref{sec:introduction}. For RQ1, we explain the physical meaning of each component in term of the transparency (Sec.~\ref{sec:Transparency}), and design some case studies for the post-hoc interpretability (Sec.~\ref{sec:Interpretability}). For RQ2, we argue that the complex-valued representation can  model different aspects of semantics and naturally address the non-linear semantic compositionality, as discussed in  Sec.~\ref{sec:complex_word_embedding}.

\begin{table}[t]

\small
\caption{Physical meanings and constraints}\label{table:corresponding}
\resizebox{0.47\textwidth}{!}{
\begin{tabular}{lll}
 \hline
  Components &  DNN &   CNM \\ \hline

Sememe &\makecell [l] { - }  &\makecell [l] {  complex basis vector / \textbf{basis state} \\ $ \{ w| w
 \in \mathcal{C}^n, ||w||_2 = 1,\} $ \\ complete~\&orthogonal  } \\
Word &\makecell [l] { real vector \\ $(-\infty, \infty)$}  &\makecell [l] { unit complex vector / \textbf{superposition state} \\ $ \{ w| w
 \in \mathcal{C}^n, ||w||_2 = 1 \}$  } \\
\makecell [l] {Low-level \\ representation} & \makecell [l] { real vector \\ $(-\infty, \infty)$}  &\makecell [l] { density matrix / \textbf{mixed system} \\ $ \{{\rho}| \rho =\rho ^*, tr(\rho) = 1 $  }\\ 
Abstraction &  \makecell [l] { CNN/RNN \\ $(-\infty, \infty)$}&  \makecell [l] { unit complex vector / \textbf{measurement} \\ $\{ w|w \in \mathcal{C}^n, ||w||_2 = 1 \}$  } \\
\makecell [l] {High-level \\representation} & \makecell [l] { real vector \\ $(-\infty, \infty)$}  &  \makecell [l] { real value/ \textbf{measured probability} \\ $(0,1)$} \\ 
 \hline
\end{tabular}}
\vspace{-5pt}
\end{table}

\begin{table}[t]
\begin{center}
\small
\caption{Selected learned important words in TREC QA. All words are lower.}\label{table:words}

\resizebox{0.48\textwidth}{!}{
\begin{tabular}{ll}
 \hline
  & Selected  words\\ \hline
Important  & \makecell[l]{studio, president, women, philosophy\\
scandinavian,  washingtonian, berliner, championship\\
defiance,  reporting, adjusted, jarred
} \\  \hline
Unimportant  & \makecell[l]{71.2,  5.5, 4m, 296036, 3.5\\
may,  be, all, born \\
movements,  economists, revenues, computers
}\\  
 \hline
\end{tabular}}
\end{center}
\vspace{-5pt}
\end{table}

\begin{table*}[t]

\begin{center}
\tiny
\caption{The matching patterns for specific sentence pairs in TREC QA. The darker the color, the bigger weight the word is. The $[$ and $]$  denotes the possible border of the current sliding windows.  }\label{tab:matching_case_study}

\resizebox{1\textwidth}{!}{
\begin{tabular}{ll}
 \hline
 Question & Correct Answer \\ \hline
Who is the [ \colorbox{blue!50}{president} \colorbox{blue!15}{or} \colorbox{blue!40}{chief} \colorbox{blue!40}{executive} \colorbox{blue!10}{of} \colorbox{blue!30}{Amtrak} ] ? & `` Long-term success ... '' said George Warrington , [ \colorbox{blue!30}{Amtrak} \colorbox{blue!15}{'s} \colorbox{blue!50}{president} \colorbox{blue!15}{and} \colorbox{blue!40}{chief} \colorbox{blue!40}{executive} ] ." \\  \hline

When [ \colorbox{blue!15}{was} \colorbox{blue!40}{Florence} \colorbox{blue!30}{Nightingale} \colorbox{blue!20}{born} ] ? &  ,"On May 12 , 1820 , the founder of modern nursing , [ \colorbox{blue!40}{Florence} \colorbox{blue!30}{Nightingale} \colorbox{blue!10}{,} \colorbox{blue!10}{was} \colorbox{blue!20}{born} ] in Florence , Italy ."\\  \hline

When [ \colorbox{blue!10}{was} \colorbox{blue!10}{the} \colorbox{blue!40}{IFC} \colorbox{blue!25}{established} ] ?&  [ \colorbox{blue!40}{IFC} \colorbox{blue!10}{was} \colorbox{blue!25}{established}  \colorbox{blue!10}{in} ] 1956 as a member of the World Bank Group .  \\  \hline

[ \colorbox{blue!10}{how} \colorbox{blue!10}{did} \colorbox{blue!40}{women} \colorbox{blue!10}{'s} \colorbox{blue!30}{role} \colorbox{blue!40}{change} \colorbox{blue!10}{during} \colorbox{blue!10}{the} \colorbox{blue!50}{war} ] & ..., the [ \colorbox{blue!30}{World} \colorbox{blue!50}{Wars} \colorbox{blue!30}{started} \colorbox{blue!10}{a} \colorbox{blue!40}{new} \colorbox{blue!40}{era} \colorbox{blue!10}{for} \colorbox{blue!40}{women} \colorbox{blue!10}{'s} ] opportunities to ....\\

[  \colorbox{blue!10}{Why} \colorbox{blue!10}{did} \colorbox{blue!10}{the} \colorbox{blue!40}{Heaven} \colorbox{blue!10}{'s} \colorbox{blue!30}{Gate} \colorbox{blue!30}{members} \colorbox{blue!30}{commit}  \colorbox{blue!40}{suicide} ] ?, &  This is not just a case of [ \colorbox{blue!30}{members} \colorbox{blue!10}{of} \colorbox{blue!10}{the} \colorbox{blue!40}{Heaven} \colorbox{blue!10}{'s} \colorbox{blue!30}{Gate} \colorbox{blue!20}{cult} \colorbox{blue!30}{committing} \colorbox{blue!40}{suicide} ] to ...  \\   \hline
\hline
\end{tabular}
}
\end{center}
\vspace{-10pt}
\end{table*}

\subsection{Transparency}
\label{sec:Transparency}
CNM aims to unify many semantic units with different granularity e.g. sememes, words, phrases (or N-gram) and document in a single complex-valued vector space, as shown in Tab.~\ref{table:corresponding}. In particular, we formulate atomic sememes as a group of complete orthogonal basis states and words as superposition states over them. A linguistic unit with larger-granularity e.g. a word phrase or a sentence is represented as a mixed system over the words (with a density matrix, i.e. a positive semi-definite matrix with unit trace).

More importantly, trainable projection measurements are adopted to extract high-level representation for a word phrase or a sentence.
Each measurement is also directly embedded in this unified Hilbert space, as a specific unit state (like words), thus making it easily understood by the neighbor words near this specific state. The corresponding trainable components in state-of-art neural network architectures, namely, kernels in CNN and cells in RNN, are represented as arbitrary real-valued without any constrains, lead to difficulty to be understood.

\subsection{Post-hoc Interpretability}
\label{sec:Interpretability}
The Post-hoc Interpretability is shown in three group of case studies, namely word weight scheme, matching pattern and discriminative semantic measurements.
\subsubsection{Word Weighting Scheme}
Tab.~\ref{table:words} shows the words selected from the top-50 most important words as well as top-50  unimportant ones. The importance of word is based on the L2-norm of its learned amplitude embedding according to Eq.~\ref{normalization}. 
It is consistent with intuition that, the important words are more about specific topics or discriminative  nouns, while the unimportant words include meaningless numbers or super-high frequency words. Note that some special form (e.g. plural form in the last row ) of words are also identified as unimportant words, since we commonly did not stem the words.

\subsubsection{Matching Pattern}

Tab.~\ref{tab:matching_case_study} shows the match schema with local sliding windows. In a local context window, we visualize the relative weights (i.e. the weights after normalized by \emph{softmax}) for each word with darkness degrees. The table illustrates that our model is capable of identifying true matched local windows of a sentence pair. Even the some words are replaced with similar forms (e.g. commit and committing in the last case) or meanings (e.g. change and new in the fourth case), it could be robust to get a relatively high matching score. From a empirical point of view, our model outperforms other models in situations where specific matching pattern are crucial to the sentence meaning, such as when two sentences share some unordered bag-of-word combinations. To some extent, it is robust up to replacement of words with similar ones in the Semantic Hilbert Space.

\subsubsection{Discriminative Semantic Measurements}

 The semantic measurements are performed through rank-one projectors \{$\ket{x} \bra{x}$\} . From a classical point of view, each projector is associated with a superposition of fundamental sememes, which is not necessarily linked to a particular word. Since the similarity metric in the Semantic Hilbert Space can be used to indicate semantic relatedness, we rely on the nearby words of the learned measurement projectors to understand what they may refer to.

Essentially, we identified the 10 most similar words to a measurement based on the cosine similarity metric. Tab.~\ref{tab:measurement_case_study} shows part of the most similar words of 5 measurements, which are randomly chosen from the total number of $k$=10 trainable measurements for the TREC QA dataset. It can be seen that the first three selected measurements were about positions, movement verbs and people's names, while the rest were about topic of history and rebellion respectively. Even though a clear explanation of the measurements is not available, we are still able to roughly understand the meaning of the measurements in the proposed data-driven approach.

\begin{table}[t]
\begin{center}
\small
\caption{Selected learned measurements for TREC QA. They were selected according to nearest words for a measurement vector in Semantic Hilbert Space.}\label{tab:measurement_case_study}
\resizebox{0.48\textwidth}{!}{
\begin{tabular}{cl}
 \hline
  & Selected neighborhood words for a measurement vector\\ \hline
1 & andes,  nagoya, inter-american, low-caste  \\  \hline
2 &  cools, injection, boiling,adrift  \\   \hline
3 &  andrews,  paul, manson, bair  \\  \hline
4 &  historically,   19th-century,  genetic, hatchback  \\  \hline 
5 &  missile,  exile, rebellion, darkness  \\  \hline

\end{tabular}}
\end{center}
\vspace{-10pt}
\end{table}

\subsection{Complex-valued Representation}
\label{sec:complex_word_embedding}
In CNM, each word is naturally embedded as a complex vector, composed of a complex phase part, a unit amplitude part and a scalar-valued length. We argue that the amplitude part  (i.e. squared root of a probabilistic weight), corresponds to the classical word embedding with the lexical meaning, while the phase part implicitly reflects the higher-level semantic aspect e.g. polarity, ambiguity or emotion. The scalar-valued length is considered as the relative weight in a mixed system. The ablation study in Sec.~\ref{sec:ablation} confirms that the complex-valued word embedding performs better than the real word embedding, which indicates that we benefit from the complex-valued embedding on the QA task.

From a mathematical point of view, complex-valued word embedding and other complex-valued components forms a new Hilbert vector space for modelling language, with a new definitions of addition and multiplication, as well as a new inner product operation. For instance, addition in the word meaning combination is defined as
\begin{equation} \label{addition}
\begin{aligned}
z =& z_1 + z_2 = r_1e^{i\theta_1} + r_2e^{i\theta_2} \\
    =& \sqrt{r_1^2 + r_2^2+2r_1r_2\cos(\theta_2-\theta_1)} \\ &\times  
e^{i\arctan \left( \frac{r_1\sin(\theta_1) + r_2\sin(\theta_2)}{r_1\cos(\theta_1)+r_2\cos(\theta_2)} \right) }
\end{aligned}
\end{equation}
where $z_1$ and $z_2$ are the values for the corresponding element for two different word vectors $\ket{w_1}$ and $\ket{w_2}$ respectively. Both the amplitudes and complex phases of $z$ are added with a  nonlinear combination of phases and amplitudes of $z_1$ and $z_2$. A classical linear addition gives $\hat{z} = r_1 + r_2$, which can be viewed as a degenerating case of the complex-valued addition with the phase information being removed ($\theta_1  = \theta_2 = 0$ in the example).

%% file: conclusion.tex
\section{Conclusions and Future Work}

Towards the interpretable matching issue, we propose two research questions to investigate the possibility of language modelling with quantum mathematical framework. To this end, we design a new framework to model all the linguistic units in a unified Hilbert space with well-defined mathematical constrains and explicit physical meaning. We implement the above framework with neural network and then demonstrate its effectiveness in question answering (QA) task. 
Due to the well-designed components, our model is advantageous with its interpretability in term of transparency and post-hoc interpretability, and also shows its potential to use complex-valued components in NLP. 

Despite the effectiveness of the current network, we would like to further explore the phase part in complex-valued word embedding to directly link to concrete semantics such as word sentiment or word position. Another possible direction is to borrow other quantum concepts to capture the interaction and non-interaction between word semantics, such as the \textit{Fock Space}~\cite{sozzo_quantum_2014} which considers both interacting and non-interacting entities in different Hilbert Spaces. Furthermore, a deeper and robust quantum-inspired neural architecture in a higher-dimension Hilbert space like~\cite{zhangpeng2018quantum} is also worth to be investigated for achieving stronger performances with better explanatory power.





%% file: acknowledge.tex
\section*{ACKNOWLEDGEMENT}

We thank Sagar Uprety, Dawei Song, and Prayag Tiwari for helpful discussions. Peng Zhang and Peter Bruza  gave us constructive comments to improve the paper. 
The GPU computing resources are partly supported by Beijing Ultrapower Software Co., Ltd and Jianquan Li.

The three authors are supported by the Quantum Access and Retrieval Theory (QUARTZ) project, which has received funding from the European Union's Horizon 2020 research and innovation programme under the Marie Sk\l{}odowska-Curie grant agreement No. 721321.